\documentclass[a4paper, 10pt, conference]{ieeeconf}      % Use this line for a4
\usepackage{FG2021}
\usepackage{comment}
\usepackage{graphicx}
\usepackage{amssymb, amsmath}
\usepackage{xcolor}
\usepackage{multirow}
\usepackage{caption}
\usepackage{subcaption}
\usepackage{mathtools}
\usepackage{cite}
\usepackage{url} 
\usepackage{hyperref}

\FGfinalcopy % *** Uncomment this line for the final submission

\IEEEoverridecommandlockouts                              % This command is only
\def\FGPaperID{****} % *** Enter the FG2021 Paper ID here

\title{\LARGE \bf
Effectiveness of Detection-based and Regression-based Approaches
for Estimating Mask-Wearing Ratio
}

\author{\parbox{16cm}{\centering
    {\large Khanh-Duy Nguyen$^1$, Huy H. Nguyen$^{2,3}$, Trung-Nghia Le$^2$,\\ Junichi Yamagishi$^{2,3}$, Isao Echizen$^{2,3,4}$} \\
    {\normalsize
    $^1$ University of Information Technology - VNUHCM, Vietnam \ \ \ \ \ \ \ \ \ 
    $^2$ National Institute of Informatics, Japan\\
    $^3$ The Graduate University for Advanced Studies, SOKENDAI, Japan \ \ \ \ \ \ \ \ \ \ 
    $^4$ The University of Tokyo, Japan\\}
    khanhnd@uit.edu.vn, \{nhhuy, ltnghia, jyamagis, iechizen\}@nii.ac.jp}
\\}

\begin{document}

%%%%%%%%%%%%%%
% COPYRIGHT NOTICE - Uncomment correct version below
%
% The notices are from the FG 2021 LOA 
%
% Active is the "Others" option - see Case #4 in the instructions posted at: http://iab-rubric.org/fg2021
%
%%%%%%%%%%%%%%

% Case #1: For papers in which all authors are employed by the US government, the copyright notice is: 
%\IEEEoverridecommandlockouts\pubid{\makebox[\columnwidth]{U.S. Government work not protected by U.S. copyright \hfill}
%\hspace{\columnsep}\makebox[\columnwidth]{ }}

% Case #2: For papers in which all authors are employed by a Crown government (UK, Canada, and Australia), the copyright notice is:
%\IEEEoverridecommandlockouts\pubid{\makebox[\columnwidth]{978-1-6654-3176-7/21/\$31.00~\copyright{}2021 Crown \hfill}
%\hspace{\columnsep}\makebox[\columnwidth]{ }}

% Case #3: For papers in which all authors are employed by the European Union, the copyright notice is:
%\IEEEoverridecommandlockouts\pubid{\makebox[\columnwidth]{978-1-6654-3176-7/21/\$31.00~\copyright{}2021 European Union \hfill}
%\hspace{\columnsep}\makebox[\columnwidth]{ }}

% Case #4: For all other papers the copyright notice is:
%\IEEEoverridecommandlockouts\pubid{\makebox[\columnwidth]{978-1-6654-3176-7/21/\$31.00~\copyright{}2021 IEEE \hfill}
%\hspace{\columnsep}\makebox[\columnwidth]{ }}

\ifFGfinal
\thispagestyle{empty}
\pagestyle{empty}
\else
\author{Anonymous FG2021 submission\\ Paper ID \FGPaperID \\}
\pagestyle{plain}
\fi
\maketitle

%%%%%%%%%%%%%%%%%%%%%%%%%%%%%%%%%%%%%%%%%%%%%%%%%%%%%%%%%%%%%%%%%%%%%%%%%%%%%%%%
\begin{abstract}
Estimating the mask-wearing ratio in public places is important as it enables health authorities to promptly analyze and implement policies. Methods for estimating the mask-wearing ratio on the basis of image analysis have been reported. However, there is still a lack of comprehensive research on both methodologies and datasets. Most recent reports straightforwardly propose estimating the ratio by applying conventional object detection and classification methods. It is feasible to use regression-based approaches to estimate the number of people wearing masks, especially for congested scenes with tiny and occluded faces, but this has not been well studied. A large-scale and well-annotated dataset is still in demand. In this paper, we present two methods for ratio estimation that leverage either a detection-based or regression-based approach. For the detection-based approach, we improved the state-of-the-art face detector, RetinaFace, used to estimate the ratio. For the regression-based approach, we fine-tuned the baseline network, CSRNet, used to estimate the density maps for masked and unmasked faces. We also present the first large-scale dataset, the ``NFM dataset,'' which contains 581,108 face annotations extracted from 18,088 video frames in 17 street-view videos\footnote{The annotations (bounding boxes and labels), and pretrained models will be released with the publication of the paper.}. Experiments demonstrated that the RetinaFace-based method has higher accuracy under various situations and that the CSRNet-based method has a shorter operation time thanks to its compactness.
\end{abstract}

%%%%%%%%%%%%%%%%%%%%%%%%%%%%%%%%%%%%%%%%%%%%%%%%%%%%%%%%%%%%%%%%%%%%%%%%%%%%%%%%
\section{INTRODUCTION}
Throughout the Covid-19 pandemic, we have seen that the use of masks has helped to prevent infection. The situation is improving, but a system to estimate the rate of mask use would be an important analysis tool for public health officials. Several studies~\cite{cats,face_mask_worker} have focused on developing a method for automatically estimating the mask-wearing ratio from images or videos, such as those captured by surveillance cameras. However, this task is very challenging due to the small face areas, severe occlusion, and cluttered background common in images and videos of congested streets. Furthermore, a large-scale and well-annotated dataset for measuring the performance of proposed methods is lacking.

The pioneering work~\cite{loey2021fighting,yadav2020deep,zhang2021novel} mostly focused on detecting masked and unmasked faces in images by using a face detector, such as Faster R-CNN or YOLO, followed by a masked/unmasked classifier. The obtained results can be further processed and used to compile statistics and issue warnings. Due to the lack of an established dataset, a small number of images crawled from the Internet were used to evaluate the performance of the proposed methods. The results demonstrated the potential application of these methods to practical systems, e.g., surveillance systems.

Unfortunately, to push the work forward, several problems need to be tackled. First, existing methods are based solely on face detection, with no consideration given to approaches for related tasks such as crowd counting. Further research comparing different approaches is necessary to identify the strengths and weaknesses of each. Second, since the images were crawled from the Internet, the faces generally had high resolution (larger than $32\times32$ pixels) and were in frontal view. They thus differed greatly from surveillance camera images in which the face areas are typically very small and unclear due to the distance between the camera and the subjects. It is thus necessary to investigate the performance of face mask estimation under real-world conditions. 

In addition to the currently utilized face detection methods, it is also feasible to use other crowd-counting approaches~\cite{sindagi2018survey} to estimate the number of people wearing masks. Like the mask-wearing ratio estimation task, the crowd-counting task has to tackle congested scenes captured by street-view cameras. Recently introduced convolutional neural network (CNN)-based crowd-counting methods are especially efficient for congested pedestrian flows thanks to the utilization of density maps. Early crowd-counting efforts~\cite{lin2001estimation,li2008estimating} have taken a detection-based approach, using face or head detectors to count the number of people, but this is computationally demanding. More recent efforts~\cite{csrnet,decidenet} have taken a regression-based approach, using a CNN to accurately and quickly predict density maps. Detection-based methods are inefficient for handling tiny and occluded objects, which are common in street-view images, while regression-based ones can tackle them effectively. Regression-based methods predict the number of people without their localization in images. Therefore, several efforts have focused on aggregate regression-based counting and localizing using an end-to-end network~\cite{composition_counting,decidenet,point_couting}.

Although the regression-based methods have achieved good performance on crowd counting, their effectiveness on mask-wearing ratio estimation has not been investigated. We have evaluated and compared detection-based and regression-based methods on their ability to estimate the mask-wearing ratio. For the detection-based approach, we used an improved RetinaFace~\cite{retinaface} face detector enhanced with a bi-directional feature pyramid network (BiFPN)~\cite{bifpn} and trained using the focal loss function~\cite{focal_loss} to effectively classify masked/unmasked faces. For the regression-based approach, we used the Congested Scene Recognition Network (CSRNet)~\cite{csrnet}, an easily trained regression network. To compare these methods, we annotated approximately 580,000 face bounding boxes extracted from about 18,000 video frames from 17 street-view videos recorded in several Japanese cities, in both daytime and nighttime, before and during the Covid-19 pandemic.

The contributions of this paper are threefold:
\begin{itemize}
\item First, we present a comparative evaluation of two approaches to estimating the mask-wearing ratio: the detection-based approach and the regression-based approach.
\item Second, we introduce a large-scale dataset of images extracted from street-view videos for use in estimating the face mask ratio. Our dataset contains 18,088 video frames with more than 580,000 face annotations. To our best knowledge, this is the first face mask dataset containing images extracted from street-view videos. 
\item Third, we present the results of comprehensive experiments to evaluate the detection- and regression-based approaches in terms of both accuracy and operation speed. Their advantages and disadvantages are also discussed.
\end{itemize}

The remainder of the paper is organized as follows. Section 2 summarizes related work. Sections 3 and 4 respectively introduce the RetinaFace- and CSRNet-based mask-wearing ratio estimation methods used, respectively, for the detection-based and regression-based approaches in the experiments. Section 5 presents our NFM dataset. The experimental results are given and discussed in Section 6. Finally, the key points are summarized and future work is mentioned in Section 7.

\section{Related Work}
\subsection{Detection-based Approach}
Methods using the detection-based approach estimate the mask-wearing ratio by detecting faces, classifying them as masked or unmasked, and tallying the number of each. Many methods have been proposed for the face detection part, such as Single Stage Headless (SSH)~\cite{ssh}, PyramidBox~\cite{pyramidbox}, and RetinaFace~\cite{retinaface}. However, traditional face detection methods face difficulties in working with faces wearing masks, especially in challenging situations (e.g., crowded areas at night and bad weather conditions).

In pioneering work, Ge \textit{et al.} created a dataset dubbed Masked Faces (MAFA)~\cite{ge2017detecting} to overcome the lack of datasets with images of masked faces. The MAFA dataset consists of 30,811 images crawled from the Internet with 35,806 masked faces (occluded by a face mask or another object). They used a locally linear embedding CNN (LLE-CNN) for detecting masked faces. Wang \textit{et al.} subsequently proposed a face attention network (FAN)~\cite{wang2017face} for leveraging context information. Experimental results on MAFA demonstrated that FAN outperforms the LLE-CNN by more than 10\% mean average precision (mAP).

Recently, with the spread of Covid-19, several efforts have been devoted to detecting face masks only. Loey \textit{et al.}~\cite{loey2021fighting} investigated the accuracy of a well-known object detector, YOLOv2~\cite{yolo9000} with a ResNet-50 backbone, for detecting only medical face masks. They collected images from two public datasets from the Kaggle community (the Medical Masks Dataset with 682 images and the Face Mask Dataset with 853 images) to create a dataset with 1415 images. The investigation showed that YOLOv2 outperformed the LLE-CNN method~\cite{ge2017detecting} (81.0\% vs 76.1\% mAP). Batagelj \textit{et al.}~\cite{batagelj2021correctly} investigated the effectiveness of off-the-shelf face detectors for masked and unmasked faces. He constructed a dataset from the MAFA and WiderFace datasets that consists of 41,934 images with 63,072 face annotations (face size at least $40\times40$ pixels). The results showed that RetinaFace achieved the highest accuracy among the evaluated detectors.

Furthermore, for computing the value of the safety impact on the community, Almalki \textit{et al.}~\cite{cats} presented a mask-wearing detection (MWD) system for estimating the percentage of people wearing a mask and the percentage of people not wearing one or wearing one incorrectly. The MWD system adds a layer at the end of the YOLOv3 detector~\cite{yolov3} to classify faces. The super-resolution CNN architecture is used to pre-process an image before it is input to the detector. For evaluation, a new MWD dataset containing 526 images was created using images from Google Images. The system detected masked/unmasked faces with 71\% mAP. Similarly, aiming to automatically detect violations of face mask-wearing and physical distancing protocols among construction workers, Razavi \textit{et al.}~\cite{face_mask_worker} created a face mask dataset containing 1853 images and used it and the Faster R-CNN~\cite{faster_rcnn} with the Inception ResNet-V2 network to detect face masks. 

\begin{figure*}[t]
\centering
\includegraphics[width=1.0\textwidth]{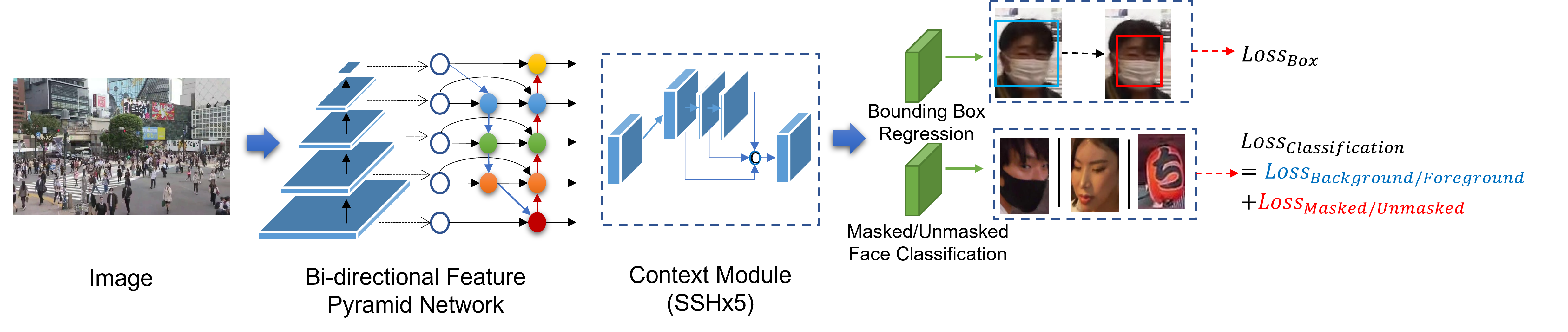}
\caption{Proposed face detection and classification pipeline based on RetinaFace.}
\label{fig:retinaface_based_pipeline}       
\end{figure*}

\begin{figure*}[t]
\centering
\includegraphics[width=1.0\textwidth]{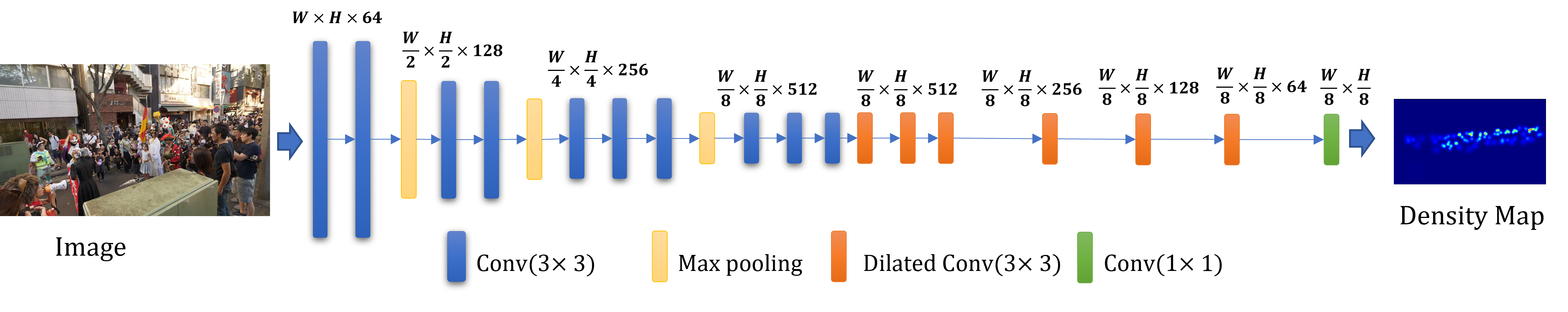}
\caption{Visualization of CSRNet architecture.} 
\label{fig:csrnet_framework}
\vspace{-5mm}
\end{figure*}

\subsection{Regression-based Approach}
Unlike the detection-based approach, methods using the regression-based approach directly predict the number of faces without detecting them. They have become mainstream methods for crowd counting thanks to the effectiveness of density maps. A CNN~\cite{csrnet,residual_regress} is typically used to predict a density map from which the count can be derived quickly. Several groups ~\cite{composition_counting,decidenet,point_couting} have recently proposed combining density map estimation and counting with detection in a unified network. Idrees \textit{et al.}~\cite{composition_counting} proved that count prediction, density maps, and detection are interrelated and can be efficiently solved by training a CNN with the proposed composition loss. Similarly, Liu \textit{et al.}~\cite{point_couting} presented a crowd-counting method that can detect human heads. This method can be trained with only point annotations.

For crowd counting, the question of which is the better approach, detection-based or regression-based, also comes. As compared by Liu \textit{et al.}~\cite{decidenet}, a detection-based method counts people accurately for low-density areas but is unreliable for congested areas. On the other hand, a regression-based method tends to overestimate the number of people in low-density areas. Gomez \textit{et al.}~\cite{gomez2021deep} compared regression- and detection-based methods for counting fruit and grains in an image. They concluded that the approaches are comparable when the density in the image is low and that regression is more accurate when the density is higher.

\section{RetinaFace-based Mask-wearing Ratio Estimation}
\subsection{Improved RetinaFace-based Detector}
The pipeline of the improved RetinaFace-based detector used for our evaluation is illustrated in Figure~\ref{fig:retinaface_based_pipeline}. It consists of three modules: a feature pyramid network for extracting features, a context module for integrating context information, and two prediction heads for bounding box regression and masked/unmasked face classification. 

\subsubsection{Bi-directional Feature Pyramid Network}
We use a BiFPN~\cite{bifpn} for extracting multi-scale features at different resolutions. The architecture of the BiFPN is illustrated in Figure~\ref{fig:retinaface_based_pipeline}. 
 
Given a list of input pyramid features $m^{\mathrm{in}} = (m^{\mathrm{in}}_{1}, m^{\mathrm{in}}_{2}, ...)$, where $m^{\mathrm{in}}_{i}$ represents the feature map at level $i$ in the pyramid, which has a resolution of $\frac{1}{2^i}$ that of the input images. The BiFPN fuses the different feature layers and then creates a list of better features: $m^{\mathrm{out}} = \mathcal{F}(m^{\mathrm{in}})$, where $\mathcal{F}$ denotes a transformation.

The conventional FPN~\cite{fpn} uses feature maps from level 3 to 7 in the input feature pyramid $m^{in} = (m^{\mathrm{in}}_3, ..., m^{\mathrm{in}}_7)$ and aggregates them in a top-down manner:
\begin{align}
m^{\mathrm{out}}_7 &= \mathrm{Conv} (m^{\mathrm{in}}_7) \\
m^{\mathrm{out}}_6 &= \mathrm{Conv}(m^{\mathrm{in}}_6 + \mathrm{Resize} (m^{\mathrm{out}}_7)) \\
\dots \nonumber \\ 
m^{\mathrm{out}}_3 &= \mathrm{Conv} (m^{\mathrm{in}}_3 + \mathrm{Resize} (m^{\mathrm{out}}_4)), 
\end{align}
where $\mathrm{Resize}(\cdot)$ is usually an upsampling or downsampling operation for resolution matching, and $\mathrm{Conv}$ is usually a convolutional operation for feature processing. 

Instead of fusion in a top-down manner, the BiFPN integrates feature layers in both directions: top-down and bottom-up. For example, feature layer $\widetilde{m}^{\mathrm{out}}_6$ is computed as
\begin{align}
m^\prime_6 &= \mathrm{Conv}\left( \frac{w_1 m^{\mathrm{in}}_6 + w_2 \mathrm{Resize} (m^{\mathrm{in}}_7)}{w_1+w_2+\epsilon} \right) \\
\widetilde{m}^{\mathrm{out}}_6 &= \mathrm{Conv} \left( \frac{w'_1 m^{\mathrm{in}}_6 + w'_2 m^\prime_6 + w'_3 \mathrm{Resize} (m^{\mathrm{out}}_5)}{w'_1+w'_2+w'_3+\epsilon} \right),
\end{align}
where $w_i$ and $w'_i$ are learnable weights.
%that can be scalars (per feature), vectors (per channel), or multi-dimensional tensors (per pixel).

\subsubsection{Context Module}
Inspired by SSH~\cite{ssh} and RetinaFace~\cite{retinaface}, we also apply independent context modules to the feature pyramid levels to increase the receptive field size and leverage the context information. The use of sequential $3\times3$ filters in the context module increases the size of the receptive field in proportion to the stride of the corresponding layer, which increases the target scale of each detection module.

\subsubsection{Prediction Heads}

\begin{comment}
\begin{figure*}
\includegraphics[width=1.0\textwidth]{Figs/DensityMapEstimation.png}
\caption{Density map estimation.}
\label{fig:DensityMapEstimation}       
\end{figure*}
\end{comment}

\paragraph{Anchors} We use anchor boxes with different sizes on feature maps, similar to their use in RetinaFace~\cite{retinaface}. There are three aspect ratios (1:2, 1:1, 2:1) for each anchor box. A length $K$ one-hot vector and a 4-coordinate vector are assigned to each anchor. The one-hot vector is the classification target, and the 4-coordinate vector is the bounding box regression target. Specifically, $K=3$, corresponding to three labels: masked, unmasked, and background.

\begin{figure}
\centering
\includegraphics[width=1.0\columnwidth]{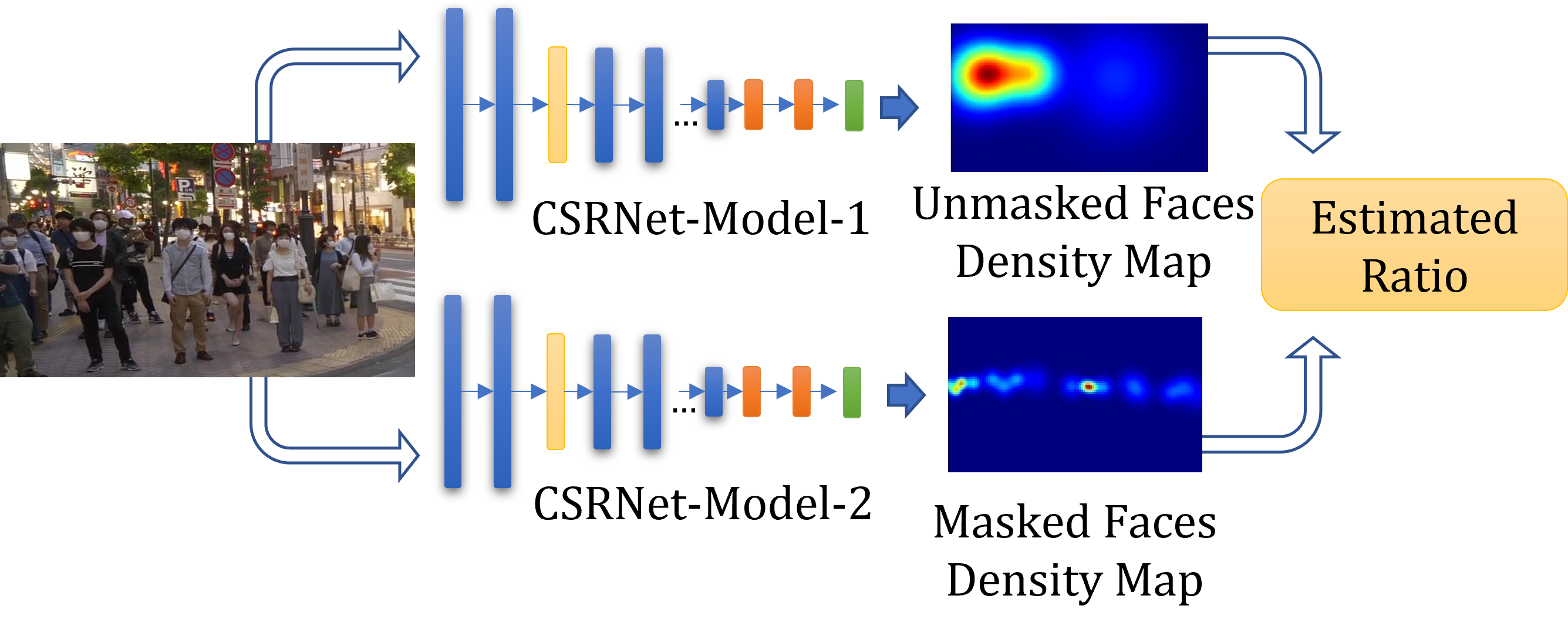}
\caption{CSRNet-based mask-wearing ratio estimation pipeline.}
\label{fig:csrnet_estimation_pipeline}
\vspace{-5mm}
\end{figure}

\paragraph{Masked/Unmasked Face Classification Head} We attached a fully convolutional network (FCN) subnet to each BiFPN level, similar to its use in RetinaFace~\cite{retinaface}, for predicting the probability of a masked/unmasked face at each anchor position. Each FCN (``face classification subnet") consists of four $3\times3$ convolutional layers, each with 256 filters, followed by ReLU activations and sigmoid activations.

\paragraph{Bounding Box Regression Head} To regress the offset from each anchor box to a nearby ground-truth face, another FCN is attached to each pyramid level in parallel with the face classification subnet. The architecture of this subnet is similar to that of the classification subnet except that it outputs four values corresponding to the relative offsets between the anchor and the ground-truth box. 

\paragraph{Multi-task Loss} For any training anchor $i$, we minimize the multi-task loss:
\begin{equation}
\mathcal{L} = \sum_{^\forall i} \mathcal{L}_{\mathrm{obj}}(p_i, p_i^*) + p_i^*\mathcal{L}_{\mathrm{cls}}(\bar{p}_i, \bar{p}_i^{*}) + p_i^*\mathcal{L}_{\mathrm{box}}(t_i, t_i^*),
\end{equation}
where $\mathcal{L}_{\mathrm{obj}}(p_i, p_i^*)$ is the binary cross entropy in which $p_i$ is the predicted probability of anchor $i$ being a face: $p^*_i$ is 1 for a positive anchor and 0 for a negative anchor. A major element of $\mathcal{L}_{\mathrm{cls}}(\bar{p}_i, \bar{p}_i^{*})$ is an improved binary cross-entropy loss, the ``focal loss"~\cite{focal_loss}, where $\bar{p}_i$ is the predicted probability of anchor $i$ being a masked face, and $\bar{p}^*_i$ is 1 for a masked face anchor and 0 for an unmasked face anchor. The focal loss addresses the imbalance ratio between the number of masked and unmasked faces. The face box regression loss is represented by $\mathcal{L}_{\mathrm{box}}(t_i, t^*_i)$, where $t_i = (t_{x_i}, t_{y_i}, t_{w_i}, t_{h_i})$, and $t^*_i = (t^*_{x_i}, t^*_{y_i}, t^*_{w_i}, t^*_{h_i})$ denote the coordinates of the predicted boxes and ground-truth ones associated with the positive anchor, respectively. 

\subsection{Ratio Estimation}
Given an input image, the improved RetinaFace-based detector detects both masked and unmasked faces. The detected faces are filtered using a confidence threshold (which was set to 0.5 in the experiments). The numbers of detected masked faces, unmasked faces, and all faces are then tallied. The mask-wearing ratio is simply calculated by dividing the number of masked faces by the total number of detected faces. 

In detail, the improved RetinaFace-based detector is trained using the stochastic gradient descent (SGD) optimizer with the momentum and weight decay at 0.9 and 0.0005, respectively. The learning rate starts at 0.01 and is divided by 10 at 50 and 68 epochs. Our training process stops after 80 epochs.

\section{CSRNet-based Mask-wearing Ratio Estimation}
\subsection{Dilated Convolutional Neural Network - CSRNet}
Using CSRNet~\cite{csrnet} is an accurate way to count by regression. CSRNet uses a conventional CNN network for extracting features followed by several dilated convolution layers for predicting a density map. The CSRNet architecture is visualized in Figure~\ref{fig:csrnet_framework}. Given an input image, the network first extracts the image features using convolutional $3\times3$ and max-pooling layers (similar to the VGG-16 architecture). It then predicts the density map by using dilated convolutional $3\times3$ and $1\times1$ layers. The resolution of the output density map is 1/8th the original resolution. We can straightforwardly apply this method to mask-wearing estimation because it produces high-quality density maps while having a pure convolutional structure.

In the training stage, a ground-truth density map is generated for each image on the basis of the annotated faces. The computation is similar to that used by CSRNet in that geometry-adaptive kernels are used to handle highly crowded scenes. Conventional density maps are computed by convolving a Gaussian kernel, which is normalized to 1, to blur the face annotations. For each face annotation, a geometry-adaptive kernel estimates the appropriate standard deviation of the Gaussian kernel by considering the distance to the $k$ nearest face annotations. The geometry-adaptive kernel is defined as
\begin{equation}
    F(x) = \sum^{N}_{i=1}\delta(x-x_i)\times G_{\sigma_i}x, \: with \: \sigma_i=\beta d_i,
\end{equation}
where $\delta(x-x_i)$ is a function representing whether there is a face at pixel $x_i$ (1 or 0), and the ground truth (face locations) of an image with $N$ labeled faces (i.e., masked and unmasked) is represented as $\sum^{N}_{i=1}\delta(x-x_i)$. Note that $\sum^{N}_{i=1}\delta(x-x_i)$ is a discrete function. A density map (a continuous function) is generated by convolving $\delta(x-x_i)$ with a Gaussian kernel with parameter $\sigma_i=\beta d_i$ (standard deviation), where $d_i$ indicates the average distance to the $k$ nearest face annotations. In our experiments, we used the configuration used by Zhang \textit{et al.}~\cite{image-crowd-counting}, with $\beta = 0.3$ and $k = 3$. 

For training CSRNet, the Euclidean distance is used to measure the distance between the estimated density map and the ground truth. The loss function is defined as
\begin{equation}
    \mathcal{L}(\Theta) = \frac{1}{2N}\sum_{i=1}^N{\lVert \mathcal{D}(X_i;\Theta) - \mathcal{D}_i^{\mathrm{GT}} \rVert}_2^2,
\end{equation}
where $X_i$ is the input image and $\Theta$ is the set of learnable parameters of CSRNet. The estimated density map of input image $X_i$ is denoted by $\mathcal{D}(X_i;\Theta)$, $\mathcal{D}_i^{\mathrm{GT}}$ is the ground-truth density map, and $N$ is the size of the training batch.

\subsection{Ratio Estimation}
To predict the mask-wearing ratio, we need to estimate the number of masked and unmasked faces. To this end, we train two CSRNet models to separately predict the numbers. After obtaining the numbers, we simply divide the number of unmasked faces by the total number of faces (masked and unmasked) to obtain the ratio. Visualization of the CSRNet-based mask-wearing ratio estimation pipeline is shown in Figure~\ref{fig:csrnet_estimation_pipeline}.

In detail, we first train a CSRNet model to estimate the number of faces. The obtained model is then fine-tuned to estimate the numbers of masked and unmasked faces separately. The SGD optimizer is applied with the momentum and weight decay set at 0.95 and 0.0005, respectively. In the experiments, we used a fixed learning rate of $\mathrm{10^{-6}}$ for the training and terminated the training after 45 epochs.

\section{NFM Dataset}
\label{sec:dataset}
\begin{table}
\caption{Details of videos obtained from Rambalac YouTube channel. 'DT': Daytime, 'NT': Nighttime, Note: Before/During Covid-19 pandemic.}
\label{table:rambalac_videos} 
\begin{center}
\begin{tabular}{c|l|l|l|l}
\hline
\textbf{ID} & \textbf{Location} & \textbf{Time Recorded} & \textbf{Type} & \textbf{Note} \\ \hline
01 & Asakusa & Aug 15, 2017 & DT+NT & Before \\
02 & Shibuya & Nov 15, 2017 & NT & Before \\
03 & Kawasaki & Jul 18, 2019 & DT & Before \\
04 & Ginza & Aug 7, 2019 & DT & Before \\
05 & Kagurazaka & Oct 18, 2019 & DT & Before \\
06 & Shibuya & Oct 31, 2019 & NT & Before \\
\hline
07 & Aomori & Jan 3, 2020 & NT+Snow & During \\
08 & Kichijoji & Feb 16, 2020 & NT & During \\
09 & Saitama & Mar 15, 2020 & DT & During \\
10 & Tokyo & Mar 18, 2020 & DT+NT+Snow & During \\
11 & Koenji & Mar 29, 2020 & DT & During \\
12 & Shibuya & Apr 19, 2020 & DT & During \\
13 & \begin{tabular}[c]{@{}l@{}}Shinjuku\\ Yotsuya\\ Ichigaya\end{tabular} & Jul 22, 2020 & DT+NT & During \\
14 & \begin{tabular}[c]{@{}l@{}}Tokorozawa\\ Aviation\end{tabular} & Jun 24, 2020 & DT & During \\
15 & \begin{tabular}[c]{@{}l@{}}Shibuya\\ Yoyogi\\ Harajuku\end{tabular} & Jun 29, 2020 & DT+NT & During \\
16 & Hachiko & Aug 19, 2020 & NT & During \\
17 & Oimachi & Aug 26, 2020 & NT & During \\ \hline
\end{tabular}
\end{center}
\vspace{-5mm}
\end{table}

\begin{figure}[t!]
\centering
\begin{subfigure}[t]{0.24\textwidth}
\centering
\includegraphics[width=\textwidth]{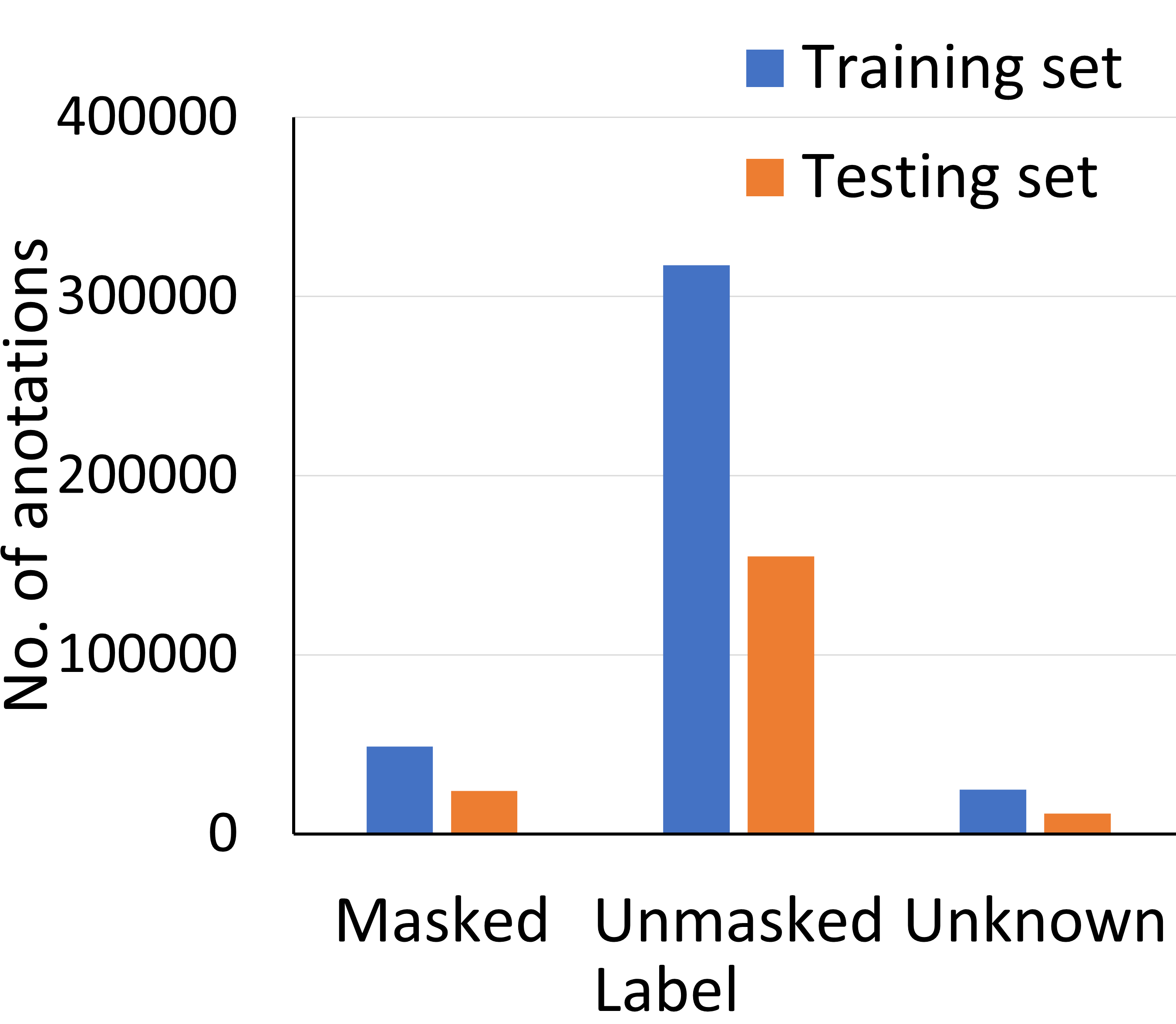}
\caption{No. of face annotations}
\end{subfigure}
~ 
\begin{subfigure}[t]{0.22\textwidth}
\centering
\includegraphics[width=\textwidth]{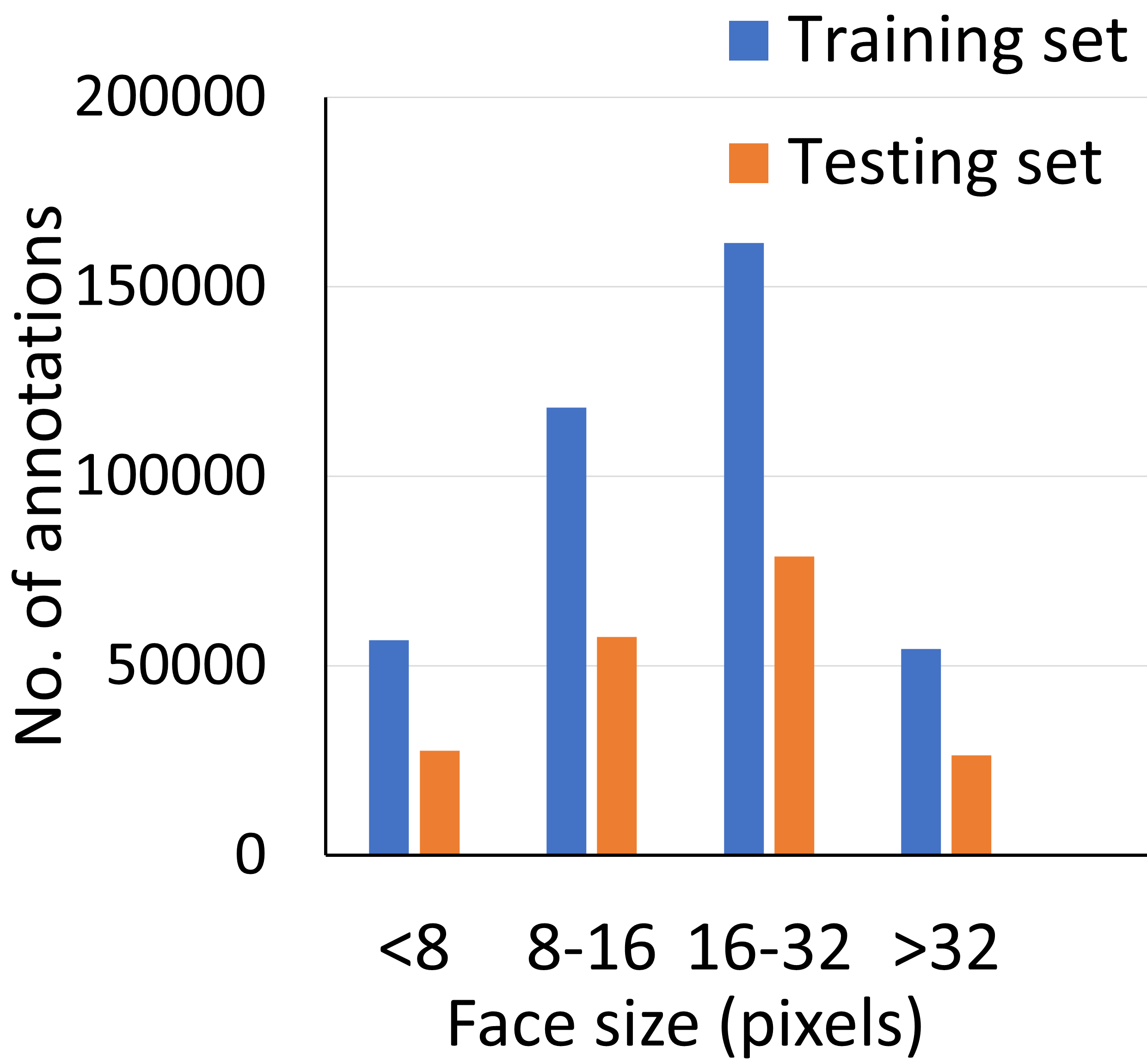}
\caption{Size of face annotations}
\end{subfigure}\\
~ 
\begin{subfigure}[t]{0.225\textwidth}
\centering
\includegraphics[width=\textwidth]{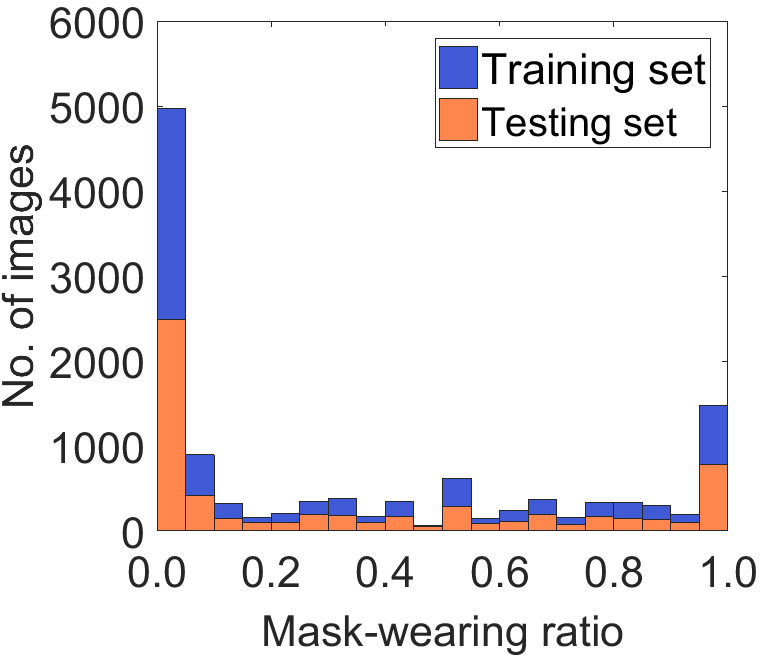}
\caption{Mask-wearing ratio}
\end{subfigure}
~
\begin{subfigure}[t]{0.225\textwidth}
\centering
\includegraphics[width=\textwidth]{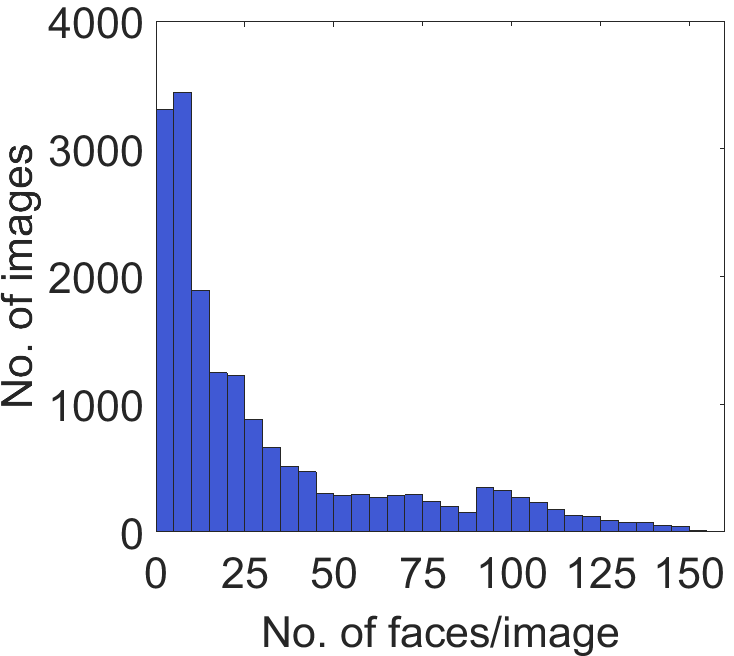}
\caption{No. of annotated faces per image}
\end{subfigure}
~
\vspace{-2mm}
\caption{Statistics on training and testing sets of NFM dataset.}
\label{fig:dataset_statistics}
\vspace{-3mm}
\end{figure}

\subsection{Dataset Creation}
We created a face mask dataset containing 581,108 face annotations extracted from 18,088 video frames in 17 street-view videos obtained from the Rambalac YouTube channel\footnote{\url{https://www.youtube.com/c/Rambalac/videos}}\footnote{The annotations (bounding boxes and labels), and pretrained models will be released with the publication of the paper.}. The details of the videos are summarized in Table~\ref{table:rambalac_videos}. The videos were taken in multiple places, at various times, before and during the Covid-19 pandemic. The total length of the videos is approximately 56 hours. As shown in the table, 6 videos were shot before the pandemic, and 11 were shot during the pandemic. The images in our dataset thus have various face mask ratios.

\begin{table}[t]
\caption{Number of annotated faces in NFM dataset.}
\label{table:dataset_number_of_label}  
\begin{center}
\begin{tabular}{c|c|ccc}
\hline
 & \textbf{Images} & \textbf{Masked} & \textbf{Unmasked} & \textbf{Unknown} \\ \hline
Training set & 12,058 & 48,736 & 317,527 & 24,594 \\ 
Testing set & \phantom{0}6,030 & 23,971 & 154,973 & 11,307 \\ %\hline
Total & 18,088 & 72,707 & 472,500 & 35,901 \\ \hline
\end{tabular}
\end{center}
\vspace{-5mm}
\end{table}

\begin{table}[t]
\caption{Average number of annotated faces per image in NFM dataset.}
\label{table:dataset_average_of_label}  
\centering
\begin{tabular}{c|ccc}
\hline
 & \textbf{Masked} & \textbf{Unmasked} & \textbf{Unknown} \\ \hline
Training set & 4.0 & 26.3 & 2.0 \\ %\hline
Testing set & 4.0 & 25.7 & 1.9 \\ \hline
\end{tabular}
\vspace{-5mm}
\end{table}

After creating our dataset, we extracted and selected frames for annotating. This process comprised three steps.
\begin{itemize}
\item Step 1 - Extract raw frames: for each video, we extracted a frame every 2 seconds. 
\item Step 2 - Detect faces: we applied the RetinaFace detector with the Resnet-50 pretrained model~\footnote{\url{https://github.com/peteryuX/retinaface-tf2}}(WiderFace dataset~\cite{widerface}) to the extracted frames to count all faces.
\item Step 3 - Select frames containing faces: we excluded raw frames containing very few face samples from our dataset, leaving us with 18,088 video frames.
\end{itemize}

\begin{figure*}[t!]
\centering
\begin{subfigure}[t]{0.43\textwidth}
\centering
\includegraphics[width=\textwidth]{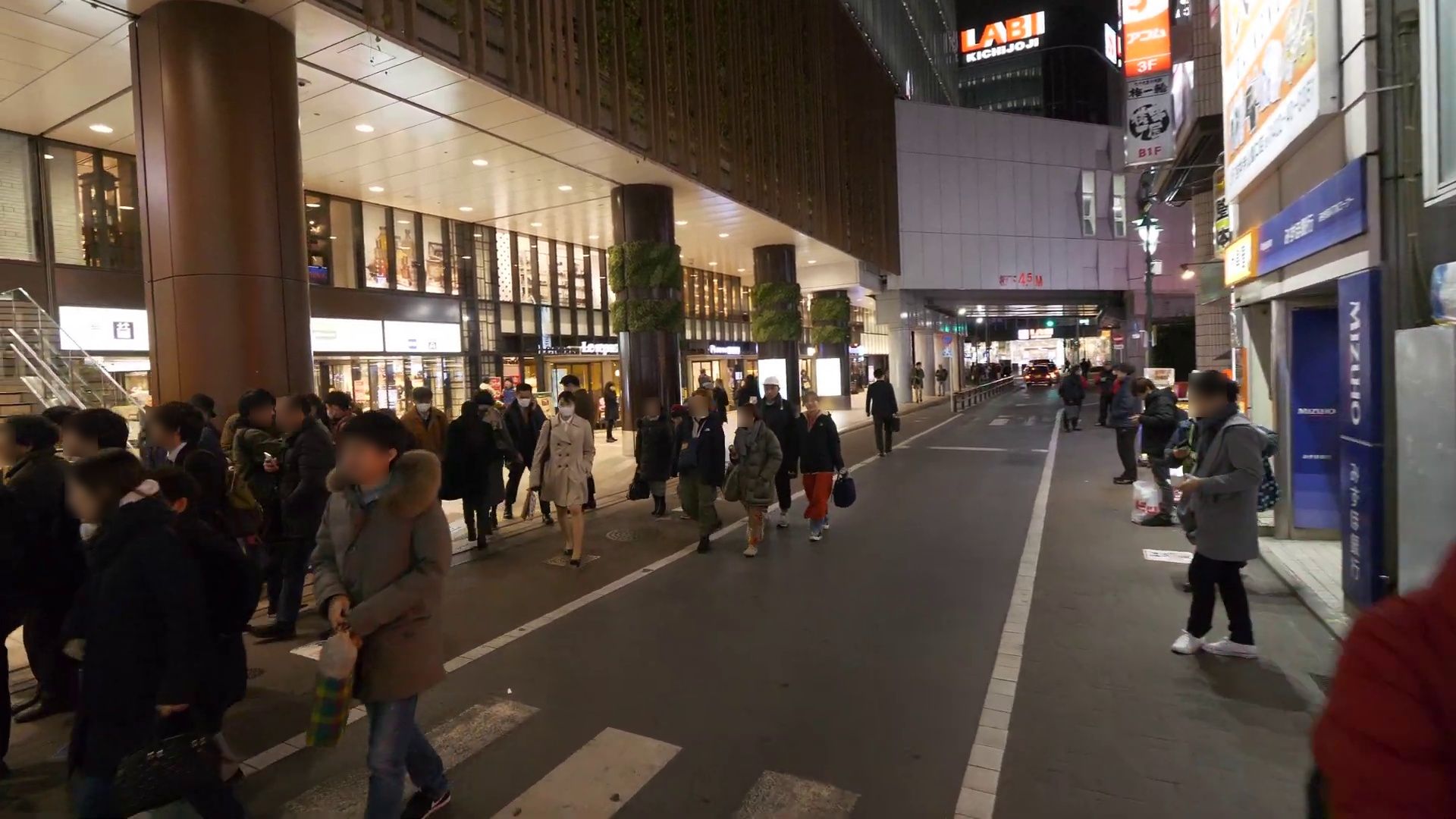}
\caption{GT: 0.19, RetinaFace: 0.23, CSRNet: 0.22, faces: 26}
\end{subfigure}%
~ 
\begin{subfigure}[t]{0.43\textwidth}
\centering
\includegraphics[width=\textwidth]{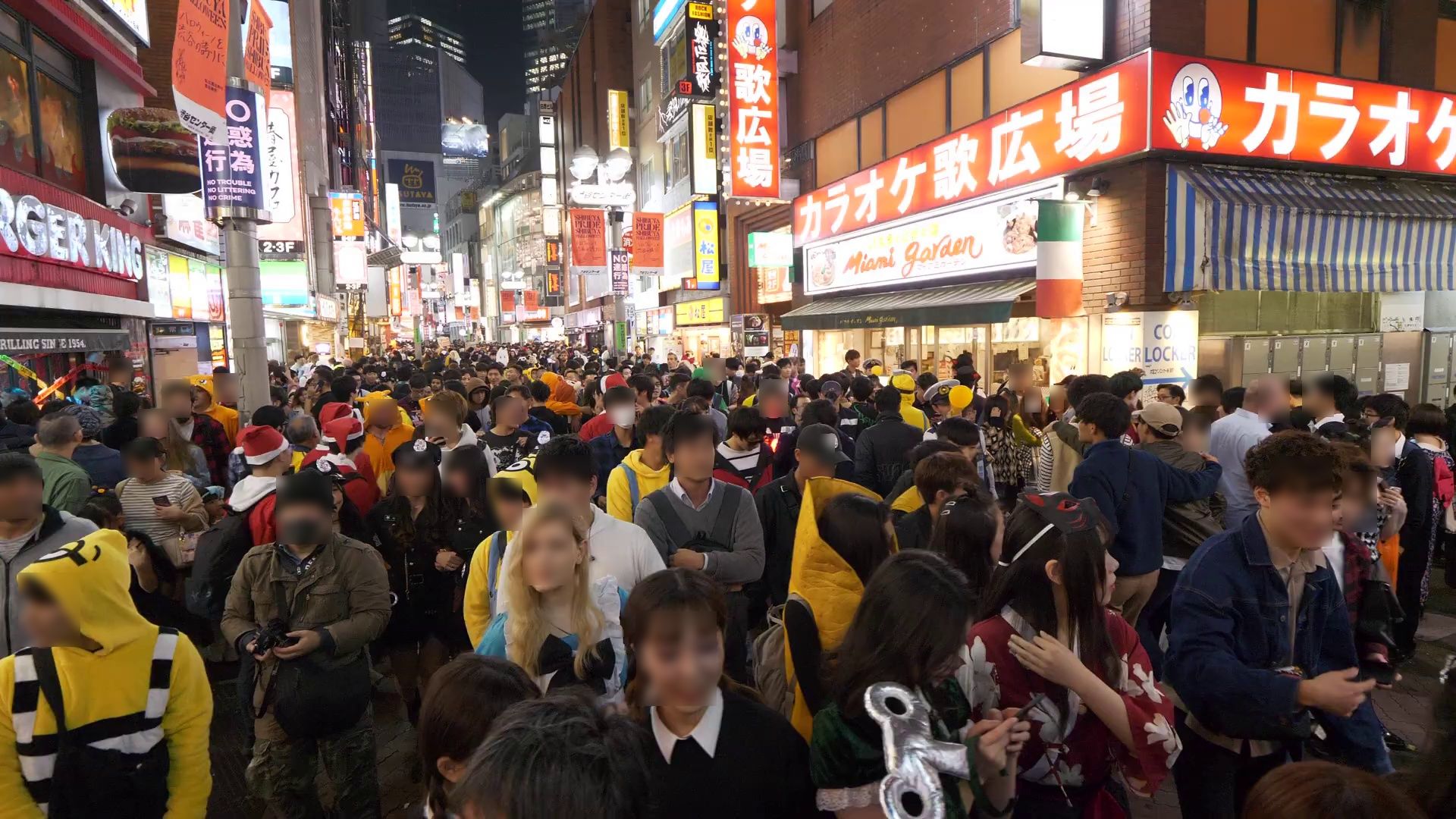}
\caption{GT: 0.06, RetinaFace: 0.04, CSRNet: 0.07, faces: 84}
\end{subfigure}
~ 
\begin{subfigure}[t]{0.43\textwidth}
\centering
\includegraphics[width=\textwidth]{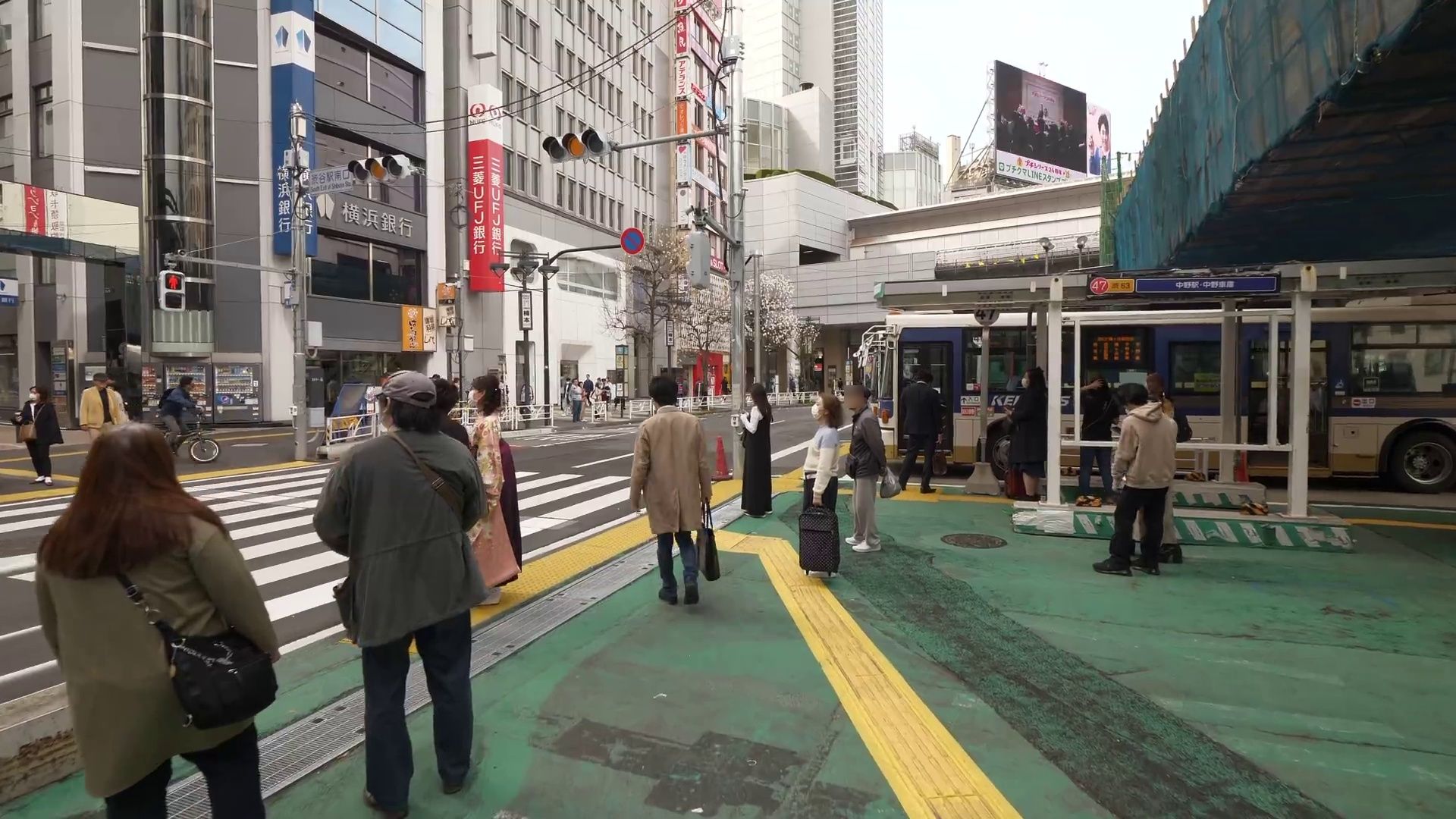}
\caption{GT: 0.55, RetinaFace: 0.60, CSRNet: 0.51, faces: 11}
\end{subfigure}
~ 
\begin{subfigure}[t]{0.43\textwidth}
\centering
\includegraphics[width=\textwidth]{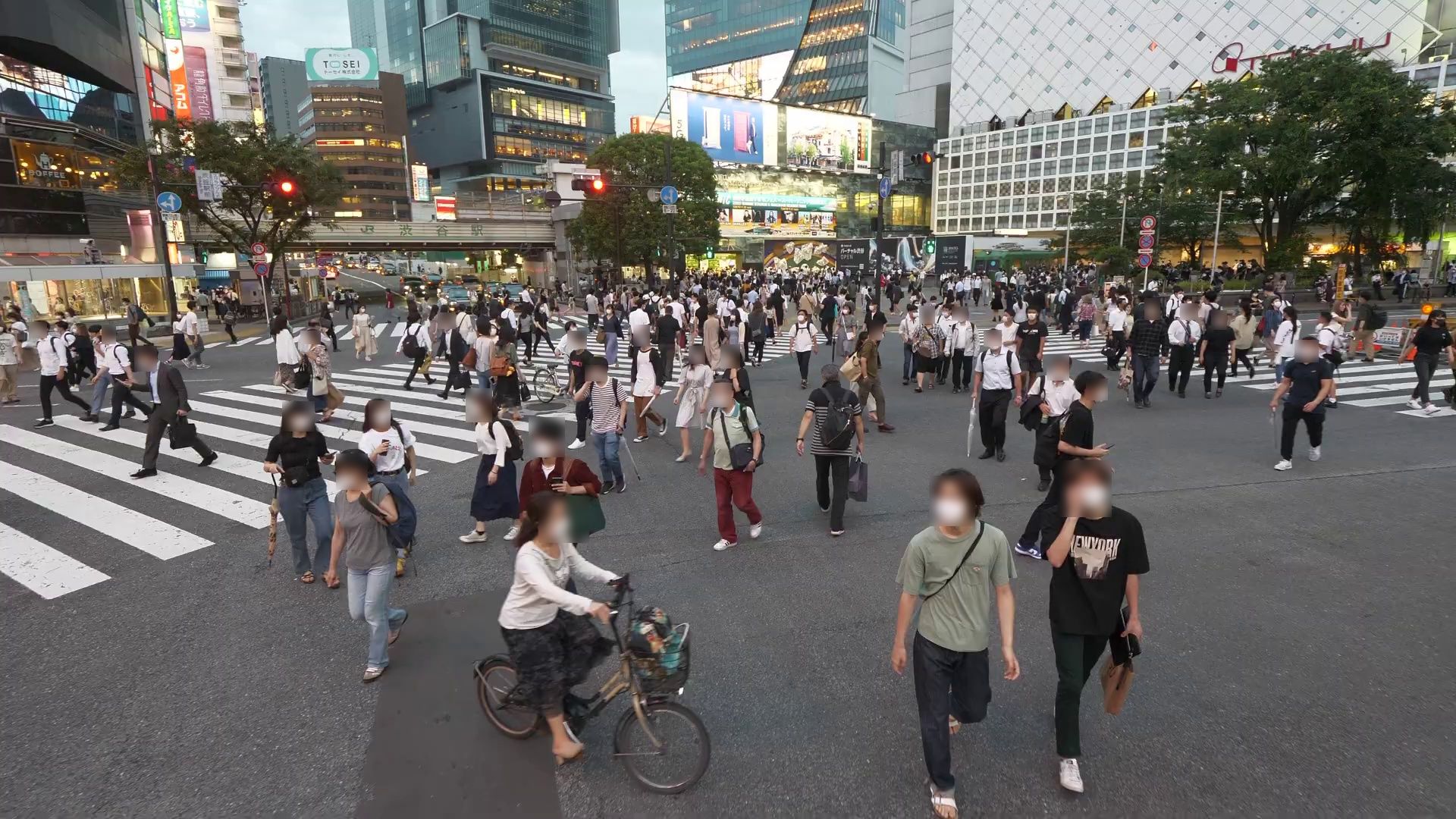}
\caption{GT: 0.93, RetinaFace: 0.95, CSRNet: 0.22, faces: 75}
\end{subfigure}
\caption{Mask-wearing ratios were estimated for four scenarios: a) a sparse scene with a low ratio; b) a dense scene with a low ratio; c) a sparse scene with a high ratio; and d) a dense scene with a high ratio. GT: ground-truth ratio. Images were extracted from videos on Rambalac channel (\url{https://www.youtube.com/c/Rambalac/videos}). Faces are blurred for anonymity.}
\label{fig:estimation_under_four_scenarios}
\vspace{-3mm}
\end{figure*}

\subsection{Image Annotation}
An image annotation comprises a bounding box and a label. Four coordinates (left, top, right, bottom) were used to denote a bounding box. The area of the face to which the bounding box was applied was the smallest square area surrounded by the hairline (upper forehead/hairline), lower jaw, and front of ears. In addition to annotating front-facing faces, we also annotated side-facing ones taken at an angle to confirm whether a mask was worn. The size of the quadrangle for an annotated face was assumed to be $10 \times 10$ pixels or more. For occluded faces, if the occlusion was judged to be more than half of the face area, annotation was not performed. For each annotated face, one of three labels was attached with a bounding box.
\begin{itemize}
\item ``Masked'': a face with a mask. The color and type of mask did not matter.
\item ``Unmasked'': a face without a mask.
\item ``Unknown'': a face for which it could not be determined whether a mask was worn due to image quality or environmental conditions.
\end{itemize}
If the mask was not properly worn, such as when the mask was stretched under the chin or hung on an ear, the “Unmasked” label was assigned. After the manual annotation, we performed verification on 20\% of the annotations (manually double-checked) to identify annotation mistakes.

\subsection{Dataset Statistics}
The statistics for our dataset are plotted in Figure~\ref{fig:dataset_statistics} and summarized in Table~\ref{table:dataset_number_of_label}. As expected, the number of masked faces is smaller than that of unmasked ones because Japanese residents were strongly encouraged to stay home during the pandemic period. On average, there are more than 30 annotated faces in each image, as shown in Table~\ref{table:dataset_average_of_label}.

\section{Experiments}
\subsection{Evaluation Metrics}
To evaluate prediction accuracy, we used the mean absolute error (MAE) and Pearson correlation metrics. The MAE metric is defined as $MAE = \frac{1}{N}\sum_{i=1}^N|c_i-c_i^{\mathrm{gt}}|$, where $N$ is the total number of images in the testing set, and $c_i$ and $c_i^{\mathrm{gt}}$ are the predicted and ground-truth counts, respectively, for image $i$. The Pearson correlation coefficient $\gamma$ is defined as 
\begin{equation}
\label{eq:equation_pearson_counting}
\gamma = \frac{\sum_{i=1}^N(c_i - \overline{c})(c_i^{\mathrm{gt}} - \overline{c}^{\mathrm{gt}})}{\sqrt{\sum_{i=1}^N(c_i - \overline{c})^2}\sqrt{\sum_{i=1}^N(c_i^{\mathrm{gt}} - \overline{c}^{\mathrm{gt}})^2}},
\end{equation}
where $\overline{c}$ and $\overline{c}^{\mathrm{gt}}$ are the mean of $c$ and $c^{\mathrm{gt}}$, respectively.

Likewise, to evaluate the mask-wearing ratio, we computed the Pearson correlation coefficient using the estimated and ground-truth ratios. 

\begin{table}[t]
\caption{Face detection results (average precision \%) for original RetinaFace detector~\cite{retinaface} and improved RetinaFace-based detector (RetinaFace$\star$) on NFM testing set.}
\label{table:face_detection_results}  
\centering
\scriptsize
\begin{tabular}{l|ccc|ccc|c}
\hline
\multirow{2}{*}{Method} & \multicolumn{3}{c|}{AP: Masked Face} & \multicolumn{3}{c|}{AP: Unmasked Face} & \multirow{2}{*}{mAP} \\ \cline{2-7}
 & L & M & S & L & M & S & \\ \hline
RetinaFace & 86.1 & 69.0 & 26.8 & 89.6 & 74.2 & 28.2 & 61.6 \\ 
RetinaFace$\star$ & \textbf{86.5} & \textbf{69.3} & \textbf{28.2} & \textbf{91.2} & \textbf{77.1} & \textbf{31.7} & \textbf{64.2} \\ \hline
\end{tabular}
\vspace{-2mm}
\end{table}

\begin{table}[t]
\caption{Face mask ratio estimation results (MAE and Pearson correlation coefficients $\gamma$) on NFM testing set. MAE (number in decimal form) is mean absolute error. FPS is number of frames processed per second.} 
\label{table:face_detection_mae_correlation}  
\centering 
\scriptsize
\begin{tabular}{l|ccc|ccc}
\hline
& \multicolumn{3}{c|}{RetinaFace$\star$} & \multicolumn{3}{c}{CSRNet} \\ \cline{2-7}
& MAE & $\gamma$ & FPS & MAE & $\gamma$ & FPS \\\hline
No. of Masked Faces & 2.41 & 0.81 & 0.81 & 3.42 & 0.38 & 6.50\\
No. of Unmasked Faces & 10.80 & 0.90 & 0.81 & 7.74 & 0.92 & 6.40\\
Total No. of Faces & 12.55 & 0.89 & 0.81 & 8.46 & 0.91 & 6.53 \\
Mask-wearing Ratio & -- & 0.94 & 0.81 & -- & 0.73 & 3.17 \\\hline 
\end{tabular}
\vspace{-6mm}
\end{table}

\subsection{Results}
We first evaluated the ability of the improved RetinaFace-based detector (RetinaFace$\star$) to detect faces. We labeled each annotated face as ``L," ``M," or ``S." ``S" was assigned to a face for which both dimensions were from 8 to 16 pixels, ``L" was assigned to a face for which both dimensions were greater than 32 pixels, and ``M" was assigned to the remaining faces. We excluded faces for which any dimension was smaller than eight pixels.

The face detection results using the conventional object detection metric—average precision (AP)—are shown in Table~\ref{table:face_detection_results}. We set the IoU (intersection over union) threshold to 0.4 because the faces in our dataset were small. The improved RetinaFace-based detector detected ``L" faces with an accuracy of 91.2\% for unmasked faces and 86.5\% for masked faces in terms of AP. The accuracy was lower for smaller faces, especially for ``S" faces, but was nevertheless higher than with the original detector. The improved RetinaFace-based detector was effective for faces larger than 16 pixels and outperformed the original detector overall by 2.6\% mAP. 

Next, we evaluated the mask-wearing ratio. To truncate the noise ratio of images containing very few faces, we set a threshold $k$ on the number of faces per image, meaning that images with fewer than $k$ faces were excluded. From observation of the images in the NFM dataset, we set $k=5$. Table~\ref{table:face_detection_mae_correlation} shows the results for the RetinaFace-based and CSRNet-based methods in terms of the MAE and correlation coefficient. The RetinaFace-based method produced good results for both metrics. The MAE scores for predicting the number of masked faces, unmasked faces, and all faces were 2.41, 10.80, and 12.55, respectively. All the estimations had correlation coefficients greater than 0.8. The correlation coefficient between the estimated mask-wearing ratio and the ground truth was 0.94.
\begin{figure}[t!]
\centering
\begin{subfigure}[t]{0.23\textwidth}
\centering
\includegraphics[width=\textwidth]{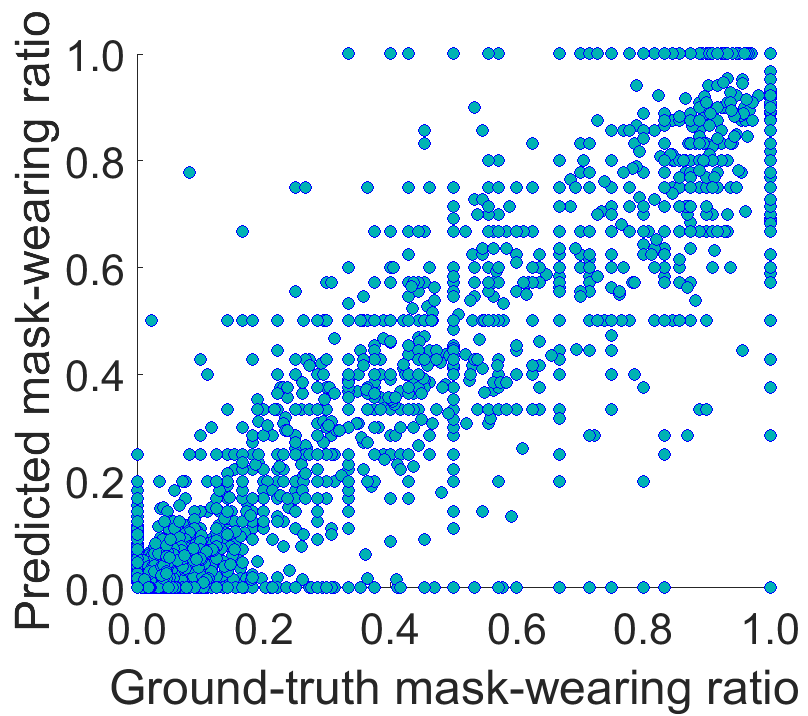}
\caption{RetinaFace$\star$}
\end{subfigure}%
~ 
\begin{subfigure}[t]{0.23\textwidth}
\centering
\includegraphics[width=\textwidth]{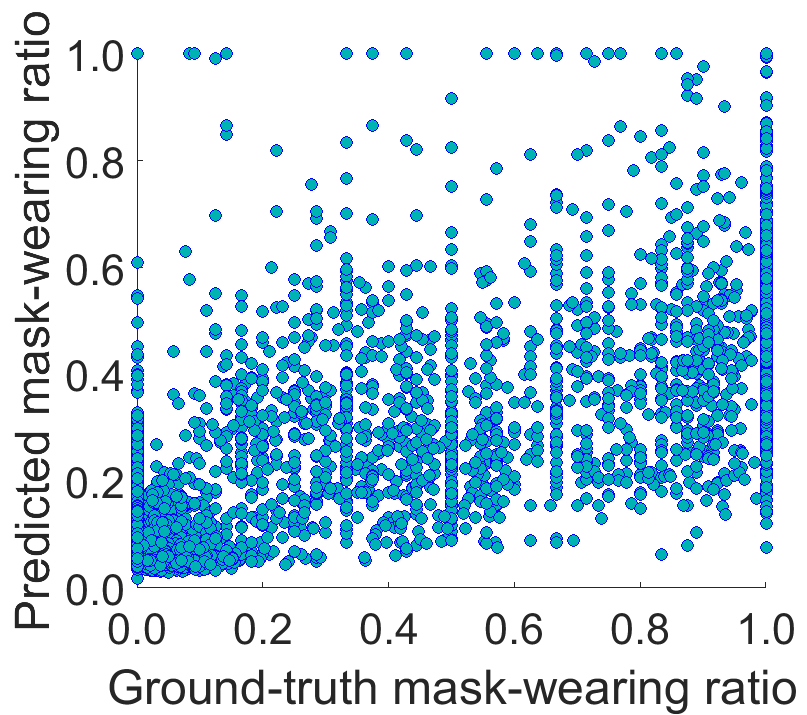}
\caption{CSRNet}
\end{subfigure}
\vspace{-2mm}
\caption{Scatter plots of predicted versus ground-truth mask-wearing ratios.}
\label{fig:ratio_estimation_visual}
\vspace{-6mm}
\end{figure}
As described above, two CSRNet models were used for the CSRNet-based method to predict the total number of faces and the number of unmasked faces. Although the CSRNet-based method estimated the total number of faces and unmasked faces with higher correlation coefficients than the RetinaFace-based one (0.91 and 0.92, respectively), the final estimated mask-wearing ratio was only 0.73. This is because CSRNet does not effectively work on classification tasks (e.g., masked/unmasked). As shown in Table~\ref{table:face_detection_results}, CSRNet did poorly on estimating the number of masked faces (0.38 correlation coefficient).

The estimation results for both methods are also shown in Figure~\ref{fig:estimation_under_four_scenarios}. The RetinaFace-based method accurately predicted the ratios for all four scenarios (sparse/dense scenes with low/high ratios) while the CSRNet-based one performed poorly for a dense scene with a high ratio. Again, this is because CSRNet does not handle the classification task effectively. It does not have a layer explicitly designed to infer the masked/unmasked probability. Furthermore, while RetinaFace leverages the FPN and SSH modules to create better features, CSRNet simply utilizes the conventional VGG-16 network, which lacks discriminative power. Hence, masked and unmasked faces cannot be distinguished, especially when the face areas are small. Visualization of the ratio estimation results for both methods on the entire testing set (Figure~\ref{fig:ratio_estimation_visual}) highlights the better performance of the improved RetinaFace-based method.

Furthermore, we computed the average mask-wearing ratio for each video in our NFM dataset to evaluate the applicability of the estimation methods used in our experiments. As shown in Table~\ref{table:ratio_estimation_videos}, the improved RetinaFace-based method produced accurate estimations for all videos—the estimated ratios are close to the actual ones. Taking a closer look at the effects of environmental conditions, we computed the accuracy of both methods on video frames extracted under daytime and nighttime conditions. As shown in Table~\ref{table:ratio_by_time}, both methods work better on video frames extracted in the daytime. This is because the camera can capture higher-quality images with clearer faces in the daytime. This enables better features to be, which enables the ratios to be estimated more precisely.
\begin{table}[t]
\caption{Average ground-truth and estimated mask-wearing ratios for each video.}
\label{table:ratio_estimation_videos}
\centering
\begin{tabular}{c|l|c|c|c|c}
\hline
\multirow{2}{*}{\textbf{ID}} & \multirow{2}{*}{\textbf{Location}} & \multirow{2}{*}{\textbf{Covid-19}} & \multirow{2}{*}{\textbf{GT Ratio}} & \multicolumn{2}{c}{\textbf{Estimated Ratio}} \\ \cline{5-6} 
 & & & & \textbf{\begin{tabular}[c]{@{}c@{}}Retina-\\ Face$\star$\end{tabular}} & \textbf{CSRNet} \\ \hline
01 & Asakusa & Before & 0.00 & 0.01 & 0.14 \\ 
02 & Shibuya & Before & 0.05 & 0.04 & 0.06 \\ 
03 & Kawasaki & Before & 0.04 & 0.02 & 0.06 \\ 
04 & Ginza & Before & 0.01 & 0.01 & 0.14 \\ 
05 & Kagurazaka & Before & 0.02 & 0.02 & 0.11 \\ 
06 & Shibuya & Before & 0.05 & 0.03 & 0.05 \\ \hline
07 & Aomori & During & 0.14 & 0.12 & 0.12 \\ 
08 & Kichijoji & During & 0.28 & 0.28 & 0.32 \\ 
09 & Saitama & During & 0.76 & 0.64 & 0.38 \\ 
10 & Tokyo & During & 0.71 & 0.58 & 0.41 \\ 
11 & Koenji & During & 0.46 & 0.42 & 0.37 \\ 
12 & Shibuya & During & 0.44 & 0.41 & 0.31 \\ 
13 & \begin{tabular}[c]{@{}l@{}}Shinjuku\\ Yotsuya\\ Ichigaya\end{tabular} & During & 0.88 & 0.83 & 0.53 \\ 
14 & \begin{tabular}[c]{@{}l@{}}Tokorozawa\\ Aviation\end{tabular} & During & 0.98 & 0.91 & 0.54 \\ 
15 & \begin{tabular}[c]{@{}l@{}}Shibuya\\ Yoyogi\\ Harajuku\end{tabular} & During & 0.88 & 0.80 & 0.42 \\ 
16 & Hachiko & During & 0.92 & 0.85 & 0.42 \\ 
17 & Oimachi & During & 0.96 & 0.88 & 0.56 \\ \hline
\end{tabular}
\vspace{-2mm}
\end{table}

\begin{table}[t]
\caption{Face mask ratio estimation results (Pearson correlation coefficients $\gamma$) on NFM testing set by time captured.}
\label{table:ratio_by_time}
\centering
\begin{tabular}{l|c|c}
\hline
\textbf{Method} & \textbf{Daytime} & \textbf{Nighttime} \\ \hline
RetinaFace$\star$ & 0.95 & 0.90 \\ 
CSRNet & 0.80 & 0.63 \\ \hline
\end{tabular}
\vspace{-6mm}
\end{table}

\subsection{Operation speed}
We used the same machine (Intel(R) Xeon(R) CPU E5-2698 v4 @ 2.20 GHz with one Tesla V-100 GPU card) to calculate the operation speed of the proposed methods. As described above and shown in Table~\ref{table:face_detection_mae_correlation}, the improved RetinaFace-based estimation method clearly outperformed the CSRNet-based one in terms of both the MAE and correlation coefficient metrics. However, it was slower. While the CSRNet-based estimation method operated at 3.17 FPS, our the RetinaFace-based estimation methodone operated at 0.81 FPS (because RetinaFace is an one-stage object detector, it predicts the bounding boxes for masked faces and unmasked faces at the same time).

An advantage of the RetinaFace-based method is its ability to accurately estimate the mask-wearing ratio while a disadvantage is its low operation speed. In contrast, the CSRNet-based one can operate four times faster but with less accuracy. Furthermore, the CSRNet models are more compact \footnote{The numbers of learnable parameters of CSRNet and improved RetinaFace are 16.26M and 29.39M, respectively.}.

\section{CONCLUSION AND FUTURE WORK}
We have presented the first comparative evaluation of detection-based and regression-based approaches for estimating the mask-wearing ratio. For detection-based estimation, we used an improved RetinaFace-based face detector enhanced with a bi-directional feature pyramid network and trained using the Focal loss function. For regression-based estimation, we used two CSRNet models to estimate the total number of faces and the number of unmask faces in video images. Evaluation of these methods on our large-scale face mask dataset (581,108 annotations) revealed the advantages and disadvantages of each approachmethod. Future work includes integrating the two approaches into a unique framework that can be jointly trained. This framework should enable efficient switching between settings to achieve accurate estimations under different conditions.

%%%%%%%%%%%%%%%%%%%%%%%%%%%%%%%%%%%%%%%%%%%%%%%%%%%%%%%%%%%%%%%%%%%%%%%%%%%%%%%%
 \section*{ACKNOWLEDGMENTS}
 This research was partly supported by JSPS KAKENHI Grants (JP16H06302, JP18H04120, JP21H04907, JP20K23355, JP21K18023), and JST CREST Grants (JPMJCR20D3, JPMJCR18A6), Japan.

%%%%%%%%%%%%%%%%%%%%%%%%%%%%%%%%%%%%%%%%%%%%%%%%%%%%%%%%%%%%%%%%%%%%%%%%%%%%%%%%

{\small
\bibliographystyle{ieeetr}
\bibliography{egbib}
}

\end{document}